\documentclass{article}

\usepackage{PRIMEarxiv}

\usepackage[utf8]{inputenc} 
\usepackage[T1]{fontenc}    
\usepackage{hyperref}       
\usepackage{url}            
\usepackage{booktabs}       
\usepackage{amsfonts}       
\usepackage{amsmath}
\usepackage{amssymb}
\usepackage{nicefrac}       
\usepackage{microtype}      
\usepackage{lipsum}
\usepackage{fancyhdr}       
\usepackage{graphicx}       
\usepackage{subfig}
\usepackage{float}
\usepackage{algorithm}
\usepackage{algpseudocode}
\usepackage{booktabs}
\usepackage{longtable}
\graphicspath{{media/}}     

\usepackage[dvipsnames]{xcolor}

\usepackage{tikz}
\usepackage{tikzpagenodes}

\usepackage{titlesec}

\title{solving a rubik's cube using its local graph structure}

\author{
  Shunyu Yao \\
  The Chinese University of Hong Kong \\
  \texttt{syyao@link.cuhk.edu.hk} \\
   \And
  Mitchy Lee \\
  Trinity University \\
  \texttt{mlee2@trinity.edu}
}

\begin{document}
\maketitle


\begin{abstract}
The Rubik’s Cube is a $3 \times 3 \times 3$ single-player combination puzzle attracting attention in the reinforcement learning community.
A Rubik’s Cube has six faces and twelve possible actions, leading to a small and unconstrained action space and a very large state space with only one goal state. 
Modeling such a large state space and storing the information of each state requires exceptional computational resources, which makes it challenging to find the shortest solution to a scrambled Rubik’s cube with limited resources.
The Rubik’s Cube can be represented as a graph, where states of the cube are nodes and actions are edges.
Drawing on graph convolutional networks, we design a new heuristic, weighted convolutional distance, for A* search algorithm to find the solution to a scrambled Rubik's Cube.
This heuristic utilizes the information of neighboring nodes and convolves them with attention-like weights, which creates a deeper search for the shortest path to the solved state.
\end{abstract}


\section{Introduction}
The Rubik's Cube is a 3-dimensional single-player combination puzzle attracting attention in the reinforcement learning community.
A $3 \times 3 \times 3$ Rubik's Cube has six faces and twelve possible actions.
These features lead to a small and unconstrained action space and a very large state space (around $4.325 \times 10^{19}$ states) with only one goal state.
Modeling such a large state space and storing the information for each state requires exceptional computational resources. 
In this situation, it is challenging to find the shortest solution to a scrambled Rubik's Cube with limited resources. \par

Previous research in the cube-solving community has developed methods to overcome this problem.
One such method, DeepCubeA, is a high performance Rubik's Cube solver developed by Agostinelli et al. that utilizes the A* search algorithm and neural networks to find a shortest path to the solved state \cite{DeepCube, DeepCubeA}. DeepCubeA trains a neural network using distance as the heuristic of the search algorithm, and yields good results on finding the shortest path to the solved cube with limited computational resources.
The success of DeepCubeA indicates that using neural networks to represent certain properties of cube states is efficient for such a large state space.\par

The Rubik's Cube can be represented as a graph, where states of the cube are nodes and actions are edges \cite{ComGroup}.
This representation can provide us with some insights from graph theory.
Graph convolutional networks (GCNs) are a powerful convolution method for handling data by using graph structures.
GCNs have a mechanism called message passing, which can pass information contained in nodes and edges to adjacent nodes, thus making use of the graph structure of the data \cite{GCN,MP1,MP2}.
Inspired by this mechanism, we propose adopting a similar method to convolve states with their distance from the solved state in order to find a solution. 
However, since the complete representation graph of the Rubik's Cube has such a large number of nodes, it is difficult to store the whole structure.
Due to limited computational power and resources, considering only the local structure of each node is a more reasonable way to utilize the information stored in the graph.\par

In this work, we propose using the A* search algorithm to find the solution to a scrambled cube with a new heuristic: weighted convolutional distance. We use the following formula to compute the weighted convolutional distance:
$$d^{(k+1)}(s) = \mu d^{(k)}(s) + (1-\mu) f_p(s)^T \mathbf{d}_{adj}^{(k)}(s)$$
where $$f_p(s) = (p_{s_R}, p_{s_r}, p_{s_L}, p_{s_l}, p_{s_U}, p_{s_u}, p_{s_D}, p_{s_d}, p_{s_F}, p_{s_f}, p_{s_B}, p_{s_b})^T$$
is a vector representing the probability of taking each of the twelve actions, $\mu \in (0,1)$ is a weight factor, $d^{(k)}(s)$ is the $k$-th convolutional weighted distance of state $s$ with $d^{(0)}(s) = f_d(s)$, $f_d$ is the distance from the solved state, and
$$\mathbf{d}_{adj}^{(k)}(s) = \left(d^{(k)} (s_R), d^{(k)} (s_r), \ldots, d^{(k)} (s_b) \right) ^T$$
is a twelve-dimensional vector with all the $k$-th convoluntional weighted distances of the adjacent states of $s$.
We use $d^{(m)}(s)$ for some $m \in \mathbb{N}$ as the heuristic of the A* search algorithm \cite{DeepCubeA,Astar1,Astar2,Astar3}.
This heuristic utilizes the information of neighboring nodes and convolves them with attention-like weights, which creates a deeper search for the shortest path to the solved state.

\section{Basic Concepts of the Rubik's Cube}
\subsection{Notations of the Rubik's Cube}
\begin{figure}[H]
    \centering
    \captionsetup[subfigure]{labelformat=empty}
    \subfloat[Action ``$U$"]{
    \begin{minipage}[b]{0.25\textwidth}
        \includegraphics[width = 1\textwidth]{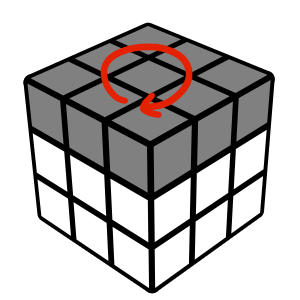}
    \end{minipage}
    \label{U}
    }
    \subfloat[Action ``$f$"]{
    \begin{minipage}[b]{0.25\textwidth}
        \includegraphics[width = 1\textwidth]{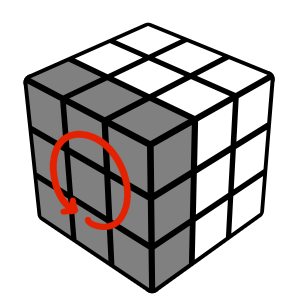}
    \end{minipage}
    \label{f}
    }
    \caption{Notations of Actions}
    \label{action_notation}
\end{figure}

The Rubik's Cube has six faces.
In this paper, we will use the following notations for faces:
\begin{figure}[H]
    \centering
    \captionsetup[subfigure]{labelformat=empty}
    \subfloat[Front]{
    \begin{minipage}[b]{0.2\textwidth}
        \includegraphics[width = 1\textwidth]{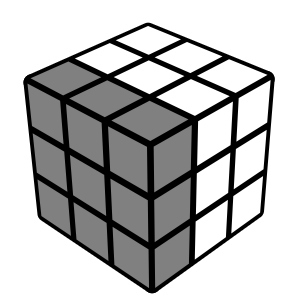}
    \end{minipage}
    \label{front}
    }
    \subfloat[Up]{
    \begin{minipage}[b]{0.2\textwidth}
        \includegraphics[width = 1\textwidth]{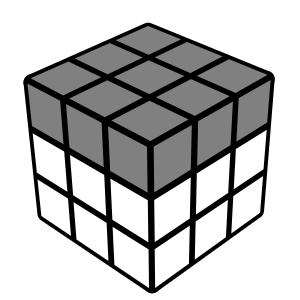}
    \end{minipage}
    \label{up}
    }
    \subfloat[Left]{
    \begin{minipage}[b]{0.2\textwidth}
        \includegraphics[width = 1\textwidth]{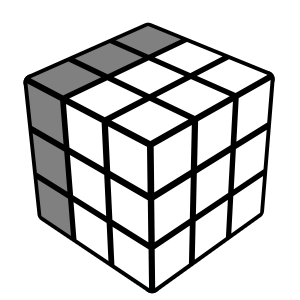}
    \end{minipage}
    \label{left}
    }

    \subfloat[Back]{
    \begin{minipage}[b]{0.2\textwidth}
        \includegraphics[width = 1\textwidth]{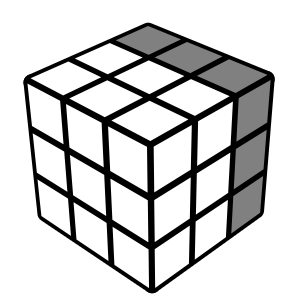}
    \end{minipage}
    \label{back}
    }
    \subfloat[Down]{
    \begin{minipage}[b]{0.2\textwidth}
        \includegraphics[width = 1\textwidth]{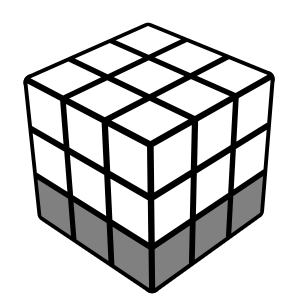}
    \end{minipage}
    \label{down}
    }
    \subfloat[Right]{
    \begin{minipage}[b]{0.2\textwidth}
        \includegraphics[width = 1\textwidth]{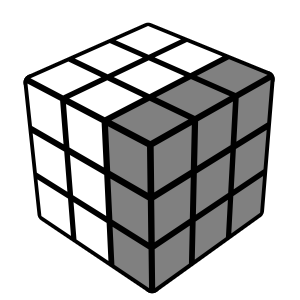}
    \end{minipage}
    \label{right}
    }
    \caption{Notations of Faces}
\end{figure}
Each face has two actions: rotating the face $90^\circ$ clockwise or $90^\circ$ counter-clockwise.
We denote clockwise actions by upper case letters and counter-clockwise actions by lower case letters, e.g., ``$U$" denotes rotating the up face by $90^\circ$ clockwise and ``$f$" denotes rotating the front face by $90^\circ$ counter-clockwise (see Figure \ref{action_notation}).
An action letter with a superscript denotes repeated action, e.g., ``$R^2$" denotes rotating the right face by $90^\circ$ clockwise twice (or $180^\circ$ clockwise). 

\subsection{Large State Space of the Rubik's Cube}
The Rubik's Cube has 26 pieces.
These pieces can be classified into three classes: fixed pieces, corner pieces, and edge pieces.
Fixed pieces (or center pieces) are the central pieces of each face, which have no influence on the states of the Rubik's Cube.
Corner pieces are the eight pieces at the corners of the cube, each with three face colors.
Edge pieces are the twelve pieces between two corner pieces, each with two face colors.

There are eight corner pieces, so there are $8!$ possible arrangements of the corner pieces. Additionally, corner pieces have three face colors, so there are three different orientations of each corner. Thus, there are $3^8$ possible orientations of the corner pieces. Although there are three possible orientations for each corner piece, only one of those orientations is correct. So, we say that $\frac{1}{3}$ of the orientations of the corner pieces are correct. Therefore, there are $8!\times 3^8\times\frac{1}{3}$ valid arrangements of the corner pieces. 

There are twelve edge pieces, so there are $12!$ possible arrangements of the edge pieces. Additionally, edge pieces have two face colors, so there are two different orientations of each edge. Thus, there are $2^{12}$ possible orientations of the edge pieces. Although there are two possible orientations for each edge piece, only one of those orientations is correct. So, we say that $\frac{1}{2}$ of the orientations of the edge pieces are correct. Therefore, there are $12!\times 2^{12}\times\frac{1}{2}$ valid arrangements of the edge pieces. 

Due to the construction of the Rubik's Cube, every permutation of the cube is even. So, of all of the possible permutations of the Rubik's Cube, $\frac{1}{2}$ of them are valid \cite{MathCube}. 

Using the above three calculations, we conclude that the total number $N$ of valid permutations of the Rubik's Cube, i.e., the total number of states in the state space, is: 
\begin{align*}
    N &= \frac{8!\times 3^8}{3}\times\frac{12!\times 2^{12}}{2}\times\frac{1}{2} \\
      &= 43252003274489860000 \\
      &\approx 4.325 \times 10^{19}.
\end{align*}

\section{Graph Convolution on the Rubik's Cube}

\subsection{Graph Representation of the Rubik's Cube}
The state space of the Rubik's Cube can be represented as a graph, where nodes are states of the cube and edges are actions \cite{ComGroup}.
The representation graph of the state space is a directed graph.
In this paper, we focus on the distance of a node from the solved state.
Since our target property is not dependent on the direction of edges, we only consider an undirected representation graph of the state space.
\begin{figure}[H]
    \begin{center}
        \begin{tikzpicture}
            \node (solved) at (0,0) {\includegraphics[width=0.15\textwidth]{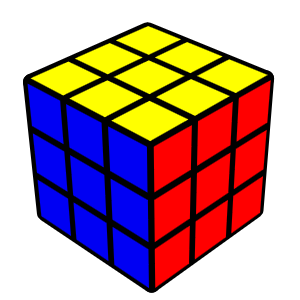}};
            \node (R) [label=above:{$R$}] at (1.15,4) {\includegraphics[width=0.1\textwidth]{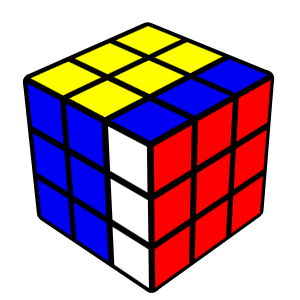}};
            \node (r) [label={[xshift=1cm, yshift=-0.5cm]$r$}] at (3,3) {\includegraphics[width=0.1\textwidth]{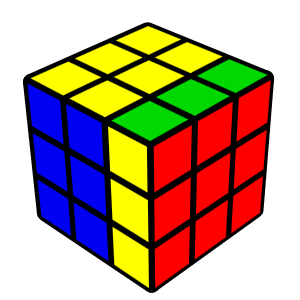}};
            \node (L) [label=right:{$L$}] at (4,1.15) {\includegraphics[width=0.1\textwidth]{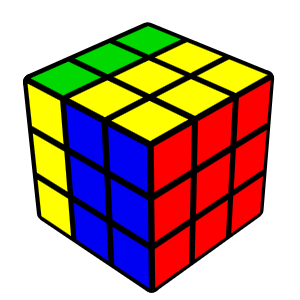}};
            \node (l) [label=right:{$l$}] at (4,-1.15) {\includegraphics[width=0.1\textwidth]{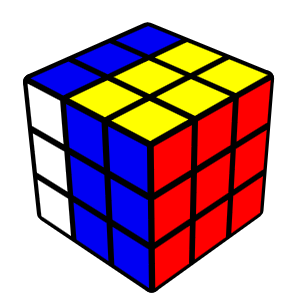}};
            \node (U) [label={[xshift=1cm, yshift=-1.9cm]$U$}] at (3,-3) {\includegraphics[width=0.1\textwidth]{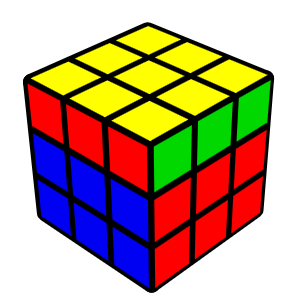}};
            \node (u) [label=below:{$u$}] at (1.15,-4) {\includegraphics[width=0.1\textwidth]{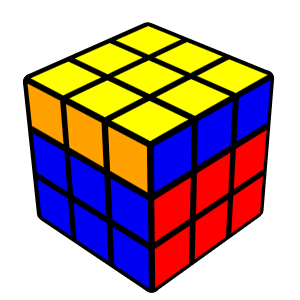}};
            \node (D) [label=below:{$D$}] at (-1.15,-4) {\includegraphics[width=0.1\textwidth]{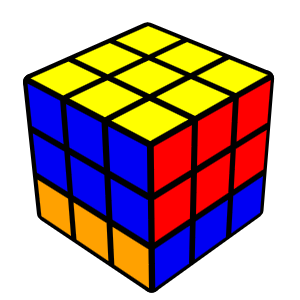}};
            \node (d) [label={[xshift=-1cm, yshift=-1.9cm]$d$}] at (-3,-3) {\includegraphics[width=0.1\textwidth]{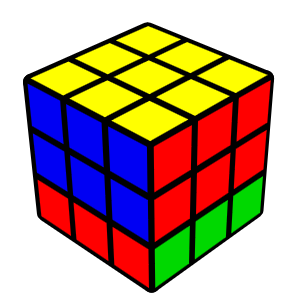}};
            \node (F) [label=left:{$F$}] at (-4,-1.15) {\includegraphics[width=0.1\textwidth]{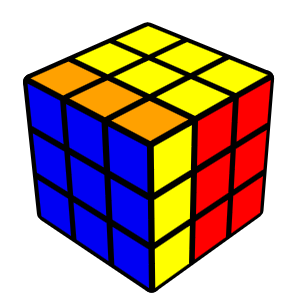}};
            \node (f) [label=left:{$f$}] at (-4,1.15) {\includegraphics[width=0.1\textwidth]{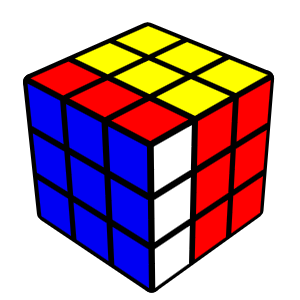}};
            \node (B) [label={[xshift=-1cm, yshift=-0.5cm]$B$}] at (-3,3) {\includegraphics[width=0.1\textwidth]{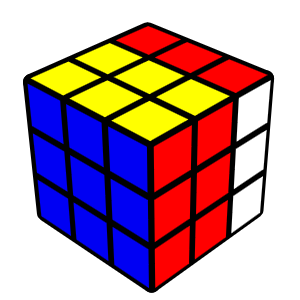}};
            \node (b) [label=above:{$b$}] at (-1.15,4) {\includegraphics[width=0.1\textwidth]{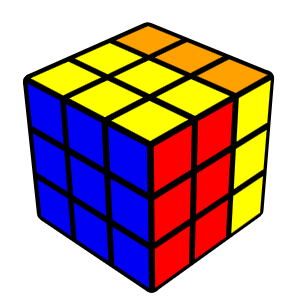}};
    
            \draw[->, line width=0.5mm] (solved) -- (R);
            \draw[->, line width=0.5mm] (1.1,1.1) -- (2.25,2.25);
            \draw[->, line width=0.5mm] (solved) -- (L);
            \draw[->, line width=0.5mm] (solved) -- (l);
            \draw[->, line width=0.5mm] (1.1,-1.1) -- (2.25,-2.25);
            \draw[->, line width=0.5mm] (solved) -- (u);
            \draw[->, line width=0.5mm] (solved) -- (D);
            \draw[->, line width=0.5mm] (-1.1,-1.1) -- (-2.25,-2.25);
            \draw[->, line width=0.5mm] (solved) -- (F);
            \draw[->, line width=0.5mm] (solved) -- (f);
            \draw[->, line width=0.5mm] (-1.1,1.1) -- (-2.25,2.25);
            \draw[->, line width=0.5mm] (solved) -- (b);
        \end{tikzpicture}
        \caption{Local graph structure of the solved state of the Rubik's Cube}
    \end{center}
\end{figure}
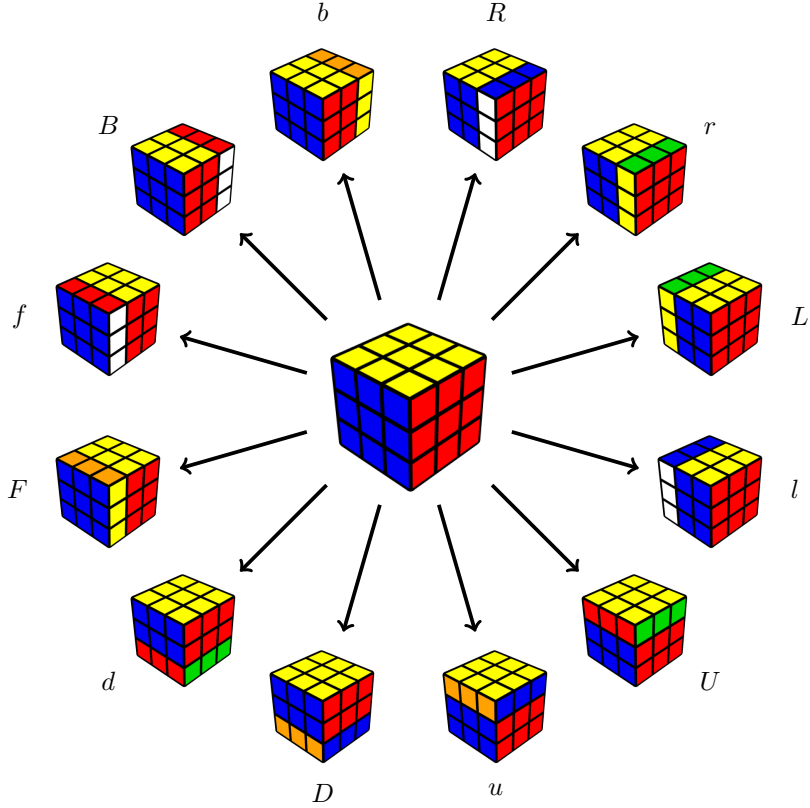

\subsection{Representation of Nodes' Property}
The Rubik's Cube has a very large state space, and thus has a very large representation graph.
Storing every possible state of the cube and computing all of their distances from the solved state requires exceptional computational power and resources. 
One way to overcome this problem is to use neural networks to represent the value of the target property \cite{DeepCube,DeepCubeA}.
We propose using two neural networks to solve the Rubik's Cube: one to compute the distance from the current state to the solved state and one to predict the next optimal move to make. \par

Distance is a crucial heuristic for searching for the shortest solution to the Rubik's Cube \cite{DeepCubeA,Astar2,Astar3}.
In this paper, we define the distance of a state of the cube as the smallest number of moves required to reach the solved state. 
We utilize the pre-trained model from the solver DeepCubeA\footnote{https://github.com/forestagostinelli/DeepCubeA} as our distance representation \cite{DeepCubeA} and denote it as $f_d$.
We also consider the next action that should be taken at each state.
The probability of each possible action offers natural attention-like weights for convolving the distance.
To obtain the probability of taking each action, we train a multilayer perceptron $f_p$\footnote{https://github.com/kongaskristjan/rubik/tree/master} with states $s$ as input and a 12-dimensional probability vector $p$ as output \cite{MLP1,MLP2}. The output vector is of the form
$$f_p(s) = (p_{s_R}, p_{s_r}, p_{s_L}, p_{s_l}, p_{s_U}, p_{s_u}, p_{s_D}, p_{s_d}, p_{s_F}, p_{s_f}, p_{s_B}, p_{s_b})^T,$$
where $p_{s_A}$ denotes the probability of taking action $A$ at the input state $s$.

\subsection{Weighted Convolutional Distance}
Graph convolutional networks are a powerful method to make predictions about the properties of graphs. 
They utilize the information stored in neighboring nodes and edges to make these predictions \cite{GCN,SGCN,SDGCN}. 
These predictions are made through a mechanism called message passing.
Message passing allows the information $i$ stored in a node to be passed to its neighboring nodes, along with a weight $w$, so that neighboring nodes can use $i$ and $w$ to update their own information accordingly \cite{MP1,MP2}. 
\begin{figure}[H]
    \centering
    \includegraphics[width=0.6\linewidth]{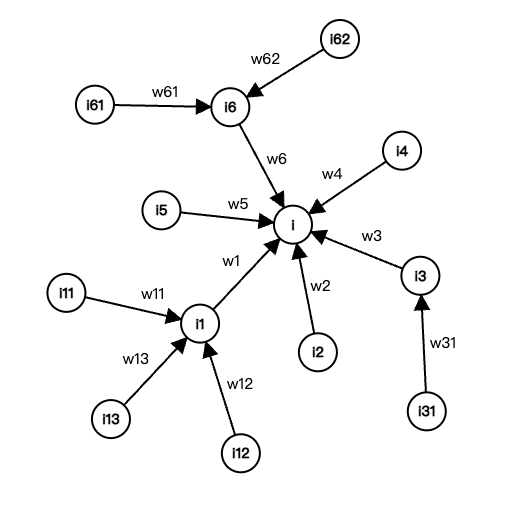}
    \caption{Message Passing}
\end{figure}
Inspired by this mechanism, we propose to convolve a state of the cube with its distance from the solved state using a similar method. 
We design our weighted convolutional distance, which uses the probability of actions as weights convolved with the distance of adjacent nodes to create deeper searching. 
The basic formula for our weighted convolutional distance is as follows:
$$d^{(1)}(s) = \mu f_d(s) + (1-\mu) \sum_{A \in Act} p_{s_A} f_d(s_A),$$
where $f_p$ is the action probability representation, $\mu \in (0,1)$ is a weight factor, $s_A$ is the state reached by taking action $A$ at the input state $s$, and $Act$ is the action space.

This formulation can be generalized to obtain a deeper convolution:
$$d^{(k+1)}(s) = \mu d^{(k)}(s) + (1-\mu) f_p(s)^T \mathbf{d}_{adj}^{(k)}(s),$$
where $f_p$ is the action representation, $d^{(k)}(s)$ is the $k$-th convolutional weighted distance of state $s$ for $k \geq 1$, and
$$\mathbf{d}_{adj}^{(k)}(s) = \left(d^{(k)} (s_L), d^{(k)} (s_l), \ldots, d^{(k)} (s_b) \right) ^T$$
is a 12-dimensional vector of the $k$-th weighted convoluntional distances of the adjacent states of $s$. 
The $k$-layer weighted convolutional distance utilizes the distance of states reached within $k$ actions from the input state, and hence represents the distance of a small neighborhood.
This will be helpful for searching since for certain bad inputs, the neural network will have a large prediction error, but that will rarely happen across the whole neighborhood. 
An important feature of the weighted convolutional distance is that it is independent of inputs, so it can be computed in parallel.
This can improve the computational efficiency in situations with many inputs.

\section{Path Search Algorithm}
\subsection{A* Search Algorithm for the Rubik's Cube}
A* search algorithm is a popular heuristic for path finding in a graph.
It classifies the nodes into two classes: closed nodes and frontier nodes. 
Closed nodes are nodes that have already been explored by the algorithm. 
They are stored in a list to prevent reprocessing nodes that have already been searched. 
Frontier nodes (also called open nodes) are nodes that have yet to be explored by the algorithm. 
They are candidates for the next step in the path. 

A* search algorithm works by closing the initial node and adding its neighboring nodes into the frontier list. 
Then  it finds the node in the frontier list with the smallest cost estimation value, adds that node into the closed list, and makes that node the working node. 
Next, the neighboring nodes of the working node are added to the frontier list, without adding any duplicates from the previous step. 
This process is repeated until the goal node is found. 
Finally, A* search algorithm will find the solution in the closed nodes and return the shortest path.

The key design of A* search algorithm is the cost estimation function \cite{Astar1,Astar2,Astar3}:
$$f(n) = g(n) + h(n).$$
In our case, $g(n)$ is the smallest number of actions need to take from the initial state $s$ to the neighboring state $n$, which can be directly computed, and
\begin{equation*}
    h(n)=
    \left\{
    \begin{aligned}
        &0 &\text{if $n$ is associated with the solved state} \\
        &d^{(m)}(n) &\text{otherwise}
    \end{aligned}
    \right.
\end{equation*}
is the $m$-layer weighted convoluntional distance, as defined above, for some $m \in \mathbb{N}$, which we use as our heuristic function \cite{DeepCube,DeepCubeA}.

\subsection{Pseudo Code}
With the cost estimation function, we propose the following algorithm for solving the Rubik's Cube:
\begin{algorithm}[H]
\caption{A* Search Algorithm for Solving the Rubik's Cube}
\begin{algorithmic}
    \State Input state $s_i$
    \State $s \gets s_i$
    \State $C \gets \{ s \}$
    \State $F \gets \{ \text{adjacent states of } s \}$
    \While{$s$ is unsolved}
        \State $n \gets \mathop{\arg\min}_{n \in F} f(n) $
        \State Add the the action $A$ taken from $s^\prime$ to $n$ to the action list $A_{s^\prime}$ of $s^\prime \in C \cap \{ \text{adjacent states of } n \}$ 
        \State $s \gets n$
        \State $C \gets C \cup \{ n \}$
        \State $F \gets (F \cup \{ \text{adjacent states of } n \}) / S$
    \EndWhile
    \While{$A_{s^\prime}$ is not empty}
        \State $s^\prime \gets s_i$
        \State Initiate the path string $P$
        \State Choose the first action $A$ in $A_{s^\prime}$
        \State Apply action $A$ to $s^\prime$
        \State Record the action $A$ to the $P$
        \If{$s^\prime$ is solved}
        \State Record the $P$ in the path list $P_l$
        \EndIf
        \If{the action list of new $s^\prime$ is empty}
        \State Apply the inverse action $A^\prime$ to $s^\prime$ and delete the action $A$ in $A_{s^\prime}$
        \EndIf
    \EndWhile
    \State $P^\prime \gets \mathop{\arg\min}_{P \in P_l} |P|$
    \State Output $P^\prime$
\end{algorithmic}
\end{algorithm}

\section{Result}
\subsection{Performance}
We test the performance of a 1-layer weighted convolutional distance (WCD) and a 2-layer WCD, and compare them with the heuristic used in DeepCubeA \cite{DeepCubeA}.
Higher layer WCDs are not included, as when the number of layers $k$ is at least three, the computation time is exceptionally long for a single input. 
To test our heuristic function, we generate 200 test samples (without duplicates), which are scrambled by performing between five and twelve random actions.
We scramble the cube only five to twelve times because A* search algorithm will take a tremendously long time to solve cubes that have been scrambled more than twelve times, due to limitations of the programming and the device we use. 
We evaluate the heuristic functions in terms of three parameters:
\begin{itemize}
    \item \textbf{Average Solution Length}: The most intuitive parameter for testing the performance of a Rubik's Cube solver.
           Better solvers should yield shorter solution lengths.
    \item \textbf{Average Time Taken}: A parameter that evaluates the time efficiency of the program. 
    \item \textbf{Average Number of Searched Nodes}: A parameter evaluating the role of the heuristic function.
           A smaller number of searched nodes means the heuristic can give a more precise direction to the searching process.            
\end{itemize}
The test result is as follows:
\begin{table}[H]
    \centering
    \begin{tabular}{c c c c}
\toprule[1pt]
    Heuristic & Length & Time (s) & No. of Searched Nodes \\
\midrule[0.5pt]
    DeepCubeA & 6.875 & 11.467 & 2938.885 \\
    1-layer WCD & 6.525 & 51.757 & 1097.345\\
    2-layer WCD & 6.455 & 326.613 & 535.000 \\
\bottomrule[1pt]
\end{tabular}
    \caption{Performance of Heuristics}
\end{table}

\subsection{Analysis}
The test result shows that our weighted convolutional distance heuristic with more convolution layers is more precise in finding a solution of the Rubik's Cube.
As the number of convolution layers increases, the average length of the solution and the number of searched nodes decreases, speaking to the efficacy of this method for finding a solution to the Rubik's Cube. 
This will decrease the memory usage required by the searching algorithm while allowing us to solve cubes that are further from the solved state. 
However, the average processing time significantly increases for deeper convolution. 
This is because that we cannot handle the convolution process in matrix form and take advantage of GPUs.
The searching efficiency will be significantly enhanced after converting to matrix form and applying fast convolution techniques in GCNs \cite{GCN,SGCN,SDGCN}.

\section{Conclusion}
Weighted convolutional distance can give a more precise search direction than DeepCubeA for A* search algorithm by convolving the distance representation with the action probability as an attention-like weight. 
It can find a shorter solution of a scrambled Rubik's Cube while searching significantly fewer nodes. 
This will decrease the memory usage while allowing us to solve cubes that are further from the solved state. \par

However, our algorithm is not very efficient, due to the inability to convert the convolution process into matrix form and apply fast convolution techniques in GCNs \cite{GCN,SGCN,SDGCN}.
Moreover, weighted convolutional distance can be generalized to similar types of combinatorial puzzles with graph-like state spaces, such as $n$-dimensional Rubik's Cubes, Sokoban, Lights Out, and sliding tile puzzles.
Our future work will focus on applying fast convolution techniques to the weighted convolutional distance and generalizing this heuristic to other combinatorial puzzles.

\bibliographystyle{unsrt}  
\bibliography{references}

\end{document}